\documentclass[11pt,a4paper]{article}
\usepackage[hyperref]{eacl2021}
\usepackage{soul}
\usepackage{times}
\usepackage{latexsym}
\usepackage{url}
\usepackage{subcaption}
\usepackage{graphicx}
\usepackage{placeins}
\usepackage{enumitem}
\usepackage{mwe}
\DeclareCaptionFont{tiny}{\tiny}

\usepackage{amsmath,amsthm,amssymb,amsfonts,mathtools}

\usepackage{tabularx}
\usepackage{multirow}
\usepackage{wrapfig}

\usepackage[justification=justified,singlelinecheck=false]{caption}

\usepackage{algorithm}
\usepackage{algpseudocode}

\usepackage{tikz-dependency}
\usepackage{tikz}
\usetikzlibrary{positioning}
\definecolor {processblue}{cmyk}{0.96,0,0,0}

\usepackage{microtype}

\urldef{\vogueurl}\url{https://www.voguebusiness.com/technology/taxonomy\%2Dis\%2Dthe\%2Dnew\%2Dfashion\%2Dtech\%2Dessential\%2Dthe\%2Dyes}
\urldef{\brainurl}\url{https://fashionbrain-project.eu/fashion\%2Dtaxonomy/}

\aclfinalcopy 
\def\aclpaperid{392} 

\title{Data Augmentation for Hypernymy Detection}

\author{Thomas Kober \\
  Rasa \\
Berlin, DE \\
  \texttt{t.kober@rasa.com} \And
  Julie Weeds \\
  University of Sussex \\
  Brighton, UK \\
  \texttt{j.e.weeds@sussex.ac.uk} \AND
  Lorenzo Bertolini \\
  University of Sussex \\
  Brighton, UK \\
  \texttt{l.bertolini@sussex.ac.uk} \And
  David Weir \\
  University of Sussex \\
  Brighton, UK \\
  \texttt{d.j.weir@sussex.ac.uk} \\\\}

\date{}

\begin{document}
\maketitle

\begin{abstract}
The automatic detection of hypernymy relationships represents a challenging problem in NLP. The successful application of state-of-the-art supervised approaches using distributed representations has generally been impeded by the limited availability of high quality training data.  We have developed two novel data augmentation techniques which generate new training examples from existing ones.  First, we combine the linguistic principles of hypernym transitivity and intersective modifier-noun composition to generate additional pairs of vectors, such as \emph{small dog - dog} or \emph{small dog - animal}, for which a hypernymy relationship can be assumed.  Second, we use generative adversarial networks (GANs) to generate pairs of vectors for which the hypernymy relation can also be assumed. We furthermore present two complementary strategies for \emph{extending} an existing dataset by leveraging linguistic resources such as WordNet. Using an evaluation across 3 different datasets for hypernymy detection and 2 different vector spaces, we demonstrate that both of the proposed automatic data augmentation and dataset extension strategies substantially improve classifier performance.
\end{abstract}

\section{Introduction}
\label{sec:introduction}
The detection of hypernymy relationships between terms represents a challenging commonsense inference problem and is a major component in recognising paraphrase and textual entailment in larger units of text.  Consequently, it is important for Question-Answering, Text Simplification and Automatic Summarization.   For example,\\[2pt]
\hspace*{10pt}\emph{There are lots of cars and vans at the port today.}\\[2pt]
might be adequately summarised by\\[2pt]
\hspace*{10pt}
\emph{There are lots of vehicles at the port today.}\\[2pt]
as \emph{car} and \emph{van} both lexically entail, i.e. they are both hyponyms of the more general term \emph{vehicle}.   

Furthermore, the recognition and discovery of hyponym-hypernym relations is a foundational part of constructing taxonomies, which has a range of practical applications in a variety of domains such as Healthcare~\citep{Barisevicius_2018} or Fashion\footnote{See e.g. \vogueurl ~or \brainurl.}.

While distributed representations of words are commonly used to find semantically similar words, they do not straightforwardly provide a way to distinguish more fine-grained semantic information, such as hypernymy, co-hyponymy and meronymy relationships. This deficiency has attracted substantial attention in the literature, and with regard to the task of hypernymy detection, both unsupervised approaches \citep{Hearst_1992,Weeds_2004,Kotlerman_2010,Santus_2014,Rimell_2014_ent,Nguyen_2017,Chang_2018_naacl} and supervised approaches \citep{Weeds_2014b,Roller_2014,Roller_2016,Shwartz_2016,Vulic_2018,Rei_2018,Kamath_2019} have been proposed.  

Supervised methods have, however, been severely hampered by a lack of adequate training data.  Not only has a paucity of labelled data been an obstacle in the adoption of deep neural networks and other more complex supervised methods, but two compounding problem-specific issues have been identified. First, there is a need to avoid lexical overlap between the training and test sets in order to avoid the lexical memorisation problem \citep{Weeds_2014b,Levy_2015}, where a supervised method simply learns the relationships between lexemes rather than generalising to their distributional features.   Second,  the performance of classifiers given just the hypernym word (at training and testing) has been shown to be almost  as good as performance given both words \citep{Weeds_2014b,Shwartz_2017}.  This suggests that classifiers are learning the distributional features that make something a more general term or a more specific term. Our conjecture is that in order to learn the more complex function, more complex machinery, and hence more labelled data is required.

In computer vision or speech recognition, it is common to use data augmentation to increase the size of the training set~\citep{Shrivastava_2017,Park_2019}.  The idea is that there are certain transformations of the data under which the class label remains invariant.  For example, rotating an image does not change whether that image contains a face or not.  By providing a supervised classifier with rotated examples, it can better generalise.  

In this work, we consider the use of linguistic transformations to augment existing datasets for hypernymy detection.  The challenge is to identify transformations that can be applied to the representations of two words that are known to be in a hypernym relationship, such that the entailment relation still holds between the transformed representations. We propose two ways to achieve this.

Our first augmentation technique is based on the hypothesis that lexical entailment is transitive and therefore invariant under certain compositions.  For example, if $A$ entails $B$ and $B$ entails $C$ then $A$ also entails $C$.  Suitable candidates for $A$ can be found by composing common intersective adjectives with the noun $B$.  For example, if we know that \emph{car} entails \emph{vehicle}, then we can augment the dataset with \emph{fast car} entails \emph{car} and \emph{fast car} entails \emph{vehicle}.

Our second augmentation technique is based on the hypothesis that lexical entailment is invariant within a certain threshold of similarity.  If $A$ entails $B$, $A'$ is very similar to $A$ and $B'$ is very similar to $B$ then $A'$ will also entail $B'$.  In order to obtain vectors which are sufficiently similar to the words in the training data, we apply generative adversarial networks (GANs) to create realistic-looking synthetic vectors, from which we choose the most similar to the words in the training data.

We evaluate the proposed techniques on three  hypernymy detection datasets.  The first two are standard benchmark tasks in this area \citep{Weeds_2014b,Baroni_2012}, both of which are generated from WordNet \citep{Fellbaum_1998}.   However, since many of the approaches to hypernmy classification involve vector space models which have been specialised using the entirety of WordNet, we need to guard against the danger that evaluations are simply measuring how well  WordNet  has been encoded, rather than how well the general hypernymy relationship has been learned.  In light of this, we  introduce a new dataset (that we call \textbf{HP4K}) which does not rely on WordNet in its construction.  

We evaluate our two data augmentation techniques against two methods for increasing the size of the training data which rely on finding or mining more non-synthetic examples.  First, we consider the extraction of additional examples from WordNet.  Second, we consider extracting examples automatically from a Wikipedia corpus using Hearst Patterns~\citep{Hearst_1992}. This provides us with what one would expect to be an upper bound on what we might reasonably expect to achieve with a similar amount of synthetic examples generated using our data augmentation techniques.

Our contributions are thus threefold.  First, we have identified two novel data augmentation techniques for the task of hypernymy detection which have the potential to generate almost limitless quantities of synthetic data.  Second, we show, rather surprisingly, that adding synthetic data is more effective than adding non-synthetic data in almost all cases.   Third, we release a new benchmark evaluation dataset for the lexical entailment task that is not dependent on WordNet.  


\section{Related Work}
\label{sec:related_work}
Data augmentation has recently become a very popular research topic in NLP and has successfully been applied in machine translation systems~\citep{Sennrich_2016,Fadaee_2017,Wang_2018_da,li-etal-2019-understanding,tong-etal-2019-supervised,matos-veliz-etal-2019-benefits,xia-etal-2019-generalized,gao-etal-2019-soft,li-specia-2019-improving,liu-etal-2019-knowledge-augmented}, but also for tasks such as relation extraction~\citep{can-etal-2019-richer,yan-etal-2019-relation}, text classification~\citep{wei-zou-2019-eda}, or natural language inference~\citep{Kang_2018,Min-etal-2020-syn}. Most similar to our usage of GANs for data augmentation is the proposal of ~\citet{Kang_2018} who leverage a GAN-based setup together with WordNet for data augmentation for natural language inference. 

While to the best of our knowledge this work represents the first application of data augmentation for lexical entailment, a number of alternative approaches have been proposed. Most proposals rely  on supervised methods for injecting an external source of knowledge into distributional representations. Starting with \textit{retro-fitting} ~\citep{Faruqui_2015b}, vector-specialization methods modify existing representations to embed desired features~\citep{Vulic_2018,Rei_2018,Kamath_2019,Glavas_2019}. 

\section{Data Augmentation Strategies for Hypernymy Detection}
\label{sec:data_augmentation}
Given a labelled dataset $\mathcal{D}_{\mathcal{X}}$ of triples $\langle x_{\emph{hypo}}^{(i)}, x_{\emph{hyper}}^{(i)}, y^{(i)} \rangle$, where $x_{hypo}^{(i)}, x_{hyper}^{(i)} \in \mathcal{X}$ and $y^{(i)} \in\{0,1\}$, we define data augmentation as adding additional hyponym-hypernym triples $\langle x^{\prime\,(j)}_{\emph{hypo}}, x^{\prime\,(j)}_{\emph{hyper}}, y^{\prime\,(j)} \rangle$ coming from an automatically generated augmentation set $\mathcal{A}_{\mathcal{X}^{\prime}}$, such that $x_{\emph{hypo}}^{\prime\,(j)}, x_{\emph{hyper}}^{\prime\,(j)} \in \mathcal{X}^{\prime}$ and $y^{\prime\,(j)} \in\{0,1\}$,  to the existing training set of $\mathcal{D}$. We ensure that the data augmentation does not introduce any lexical overlap with the existing test set, i.e. $\mathcal{X}\cap \mathcal{X}^{\prime}=\emptyset$.

Data augmentation strategies in NLP can roughly be divided into two categories: linguistically grounded augmentation and artificial augmentation. In the former, which has been the dominant paradigm in NLP, any additional instances that are added to a training set have an actual surface form representation, i.e. the data points correspond to actual words or sentences~\citep{kim-etal-2019-data,kumar-etal-2019-closer,gao-etal-2019-soft,Andreas_2020,Croce-etal-2020-ganbert}. The latter adds instances that are fully or partly artificial, meaning they do not correspond to any words or sentences. In this work we propose methods for both categories, data augmentation via distributional composition adds data points grounded in real language to a training set, and data augmentation based on GANs infers plausible points in latent space, which however, do not correspond to any real linguistic objects. 

Furthermore, we distinguish between \emph{data augmentation} and \emph{dataset extension}, where in the former case we only leverage knowledge from the existing dataset and in the latter case we rely on expanding the training set with additionally mined hyponym-hypernym pairs. Below, we discuss two  ways of augmenting and two ways of extending a training set. We make use of a cleaned October 2013 Wikipedia dump~\citep{Wilson_2015} as reference corpus to determine word and bigram frequencies.

\textbf{Distributional Composition based Augmentation.} We take a modified noun as being in a hypernymy relation with the unmodified noun. For example, we treat the pairs $\langle\mbox{\em fast car\/},\mbox{\em car\/}\rangle$ and $\langle\mbox{\em car\/},\mbox{\em vehicle\/}\rangle$ as expressing the same semantic relation when the modifier-noun compound is composed with an intersective composition function. 

We focus on adjective-noun (AN) and noun-noun (NN) compounds, extracted from our reference corpus where each AN or NN phrase occurred at least 50 times. We filtered pairs with non-subsective adjectives using a wordlist from \citet{Nayak_2014}\footnote{In preliminary experiments we did not find that filtering non-subsective adjectives had much of an effect, but decided to move forward with the filtered data nonetheless.}.  

We consider two strategies for automatically constructing positive hyponym-hypernym pairs: simple positive cases such as $\langle\mbox{\em small dog\/},\mbox{\em dog\/}\rangle$ or $\langle\mbox{\em fast car\/},\mbox{\em car}\rangle$; and gapped positive cases that mimic the transitivity of hypernym relations, where we pair the hypernym of an existing hyponym-hypernym pair with a compound hyponym. For example if $\langle\mbox{\em car\/},\mbox{\em vehicle\/}\rangle$ is in the training data, we combine \emph{car} with one of its modifiers to create the pair $\langle\mbox{\em fast car\/},\mbox{\em vehicle\/}\rangle$.

We construct negative pairs from the simple positive cases using two strategies: creating compositional co-hyponyms such as $\langle\mbox{\em fast car\/},\mbox{\em red car\/}\rangle$, where we keep the head noun fixed and pair it with two different modifiers; and creating perturbed simple positive examples, such as $\langle\mbox{\em small dog\/},\mbox{\em cat\/}\rangle$ where we select the incorrect hypernym (e.g. \emph{cat})  from the $n$ most similar nouns to the composed hyponym (e.g. \emph{dog}). We apply the same methodology to the perturbed gapped positive examples, replacing the correct hypernym with a noun from  the top $n$ neighbours of the compositional hyponym's head noun. For example, given a positive pair such as $\langle\mbox{\em dog\/},\mbox{\em animal\/}\rangle$, this would result in negative examples such as $\langle\mbox{\em small dog\/},\mbox{\em vehicle\/}\rangle$, where the hyponym \emph{dog} is paired with a modifier and the hypernym \emph{animal} is replaced with one of its neighbours, in this case, \emph{vehicle}.

In neural word embeddings, an additive composition function approximates the intersection of the corresponding feature spaces~\citep{Tian_2017}, hence by creating positive pairs such as $\langle\mbox{\em small dog\/},\mbox{\em dog\/}\rangle$, we encode the distributional inclusion hypothesis~\citep{Weeds_2004,Geffet_2005a} in the augmentation set.

\textbf{GAN based Augmentation.} We create an augmentation set using Generative Adversarial Networks~\citep{Goodfellow_2014}. GANs consist of two model components --- the generator and the discriminator --- which are typically implemented as neural networks. The generator's task is to create data that mimics the distribution of the original data, while the discriminator's task is to distinguish between data coming from the real distribution and synthetic data coming from the generator. Both components are trained jointly until the generator succeeds in creating realistic data. Using GANs for data augmentation has  been shown to be a successful strategy for a number of computer vision tasks~\citep{Shrivastava_2017,Frid_Adar_2018,Neff_2018}. Our goal is to create synthetic hyponym-hypernym pairs that are similar to real examples. Unlike most other scenarios involving GANs for NLP tasks, our generated vectors do not need to correspond to actual words.

For our model - \emph{GANDALF}\footnote{\textbf{GAN}-based \textbf{D}ata \textbf{A}ugmentation for \textbf{L}exical in\textbf{F}erence.} - we used a list of $\approx$40K nouns for which we had vector representations as the ``real" data input to \emph{GANDALF}, and sampled the synthetic vectors from a Gaussian distribution, optimising a binary cross-entropy error criterion for the generator and the discriminator, which are both simple feedforward networks with a single hidden layer. We provide \emph{GANDALF}'s full model details in Appendix~\ref{sec:supplemental_a}. As an additional quality check for the generated vectors, we tested whether a logistic regression classifier could distinguish the synthetic and non-synthetic vectors. Typically, the accuracy of the classifier was between 0.55-0.65, meaning the classifier is barely able to distinguish between ``real" vectors and generated ones.

\begin{figure*}[!htb]
\centering
\includegraphics[width=\textwidth]{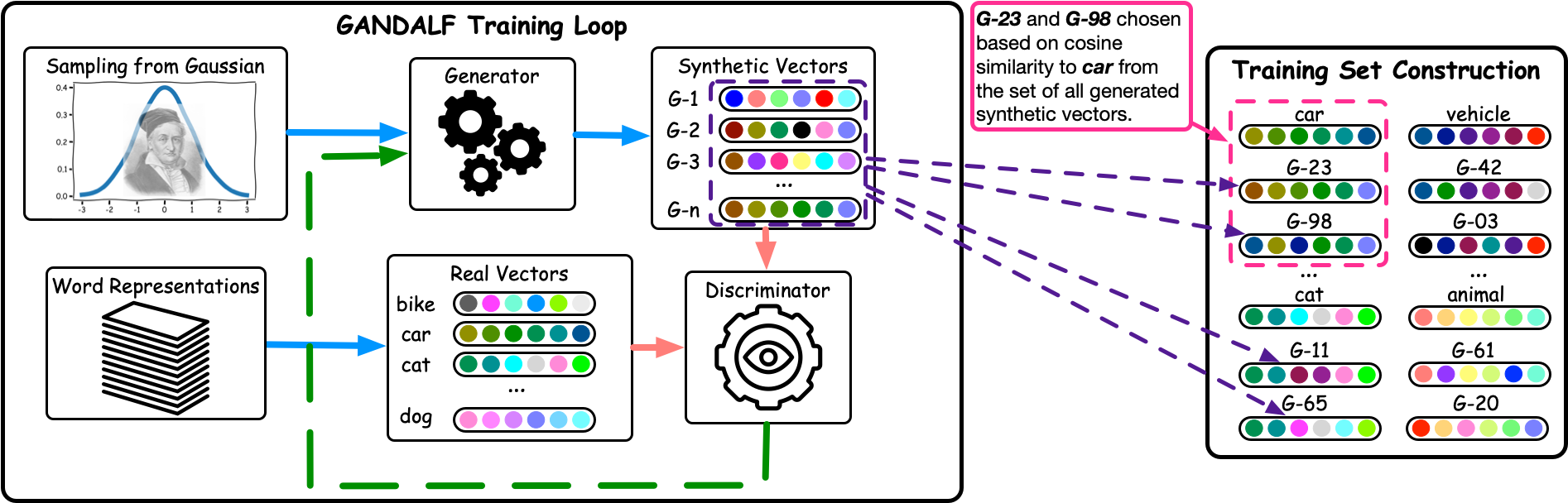}
\captionsetup{font=small}
\caption{The left panel shows the training loop for \emph{GANDALF} where synthetic noun vectors are generated from Gaussian noise. The discriminator's task is to distinguish between the generated synthetic vectors and real word representations from word2vec or hypervec. The dashed line indicates error backpropagation, on the basis of the discriminator's current performance, through to the generator. After a final set of synthetic vectors has been generated with \emph{GANDALF}, we choose top \emph{n} most similar synthetic vectors to a real representation and add these with the same label as the original hyponym-hypernym pair (e.g. $\langle\mbox{\em cat\/},\mbox{\em animal\/}\rangle$) to the training set (right panel).}
\label{fig:gan_loop}
\end{figure*}

Figure~\ref{fig:gan_loop} illustrates the training loop of \emph{GANDALF} as well as the selection process for constructing an augmented training set. Essentially, once \emph{GANDALF} has been trained, the generator is used to create a large collection\footnote{We would typically create half a million synthetic nouns.} of synthetic noun vectors. To augment a dataset, $\mathcal{D}_{\mathcal{X}}$, for each triple, $\langle x_{\emph{hypo}}, x_{\emph{hyper}}, y \rangle\in\mathcal{D}_{\mathcal{X}}$ we find the $n$ synthetic vectors most similar to $x_{\emph{hypo}}$ and the $n$ synthetic vectors most similar to $x_{\emph{hyper}}$ and for each of the $n^2$ synthetic vector pairs, $\langle x^{\prime}_{\emph{hypo}}, x^{\prime}_{\emph{hyper}}\rangle$, we create the triple $\langle x^{\prime}_{\emph{hypo}}, x^{\prime}_{\emph{hyper}}, y \rangle$. The augmented training set is formed by randomly sub-sampling this set of  triples.

\textbf{WordNet based Extension.} WordNet~\citep{Fellbaum_1998} is a large manually curated lexical resource, covering a wide range of lexical relations between words, where groups of semantically similar words form ``synsets"\footnote{We use the API provided by NLTK~\citep{Loper_2002}, using WordNet 3.0.}. For each synset we extract all hypernyms and hyponyms of a given lexeme, and add it as a positive hyponym-hypernym pair if the original lexeme and any extracted hypernym/hyponym occurs at least 30 times in our reference corpus. 

We construct negative pairs based on distributional similarity, where we calculate the pairwise cosine similarities between all lexemes in the positive set. Subsequently we use all antecedent (LHS) lexemes from the extracted positive pairs and select the top $n$ most similar words for each antecedent as negative examples\footnote{Ensuring we don't accidentally add any real positive pairs.}.   

\textbf{Pattern based Extension.} Hearst Patterns~\citep{Hearst_1992} are textual patterns such as \emph{a car \textbf{is-a} vehicle} and can be automatically mined from text corpora in an unsupervised way. This has recently been shown to deliver strong performance on the hypernymy detection task~\citep{Roller_2018}. In this work, we leverage Hearst Patterns to mine additional hyponym-hypernym pairs in order to extend a  training set. We treat any extracted noun pairs as additional positive examples and create the negative pairs in the same way as for the WordNet-based approach above.  

\section{Experiments}
\label{sec:experiments}

We evaluate  our models on the datasets \textbf{Weeds}~\cite{Weeds_2014b} and \textbf{LEDS}~\citep{Baroni_2012}: well-studied and frequently used benchmarks for the hypernymy detection task~\citep{Roller_2014,Vilnis_2014,Roller_2016,Carmona_2017,Shwartz_2017}.  Since both datasets use WordNet during construction, this can give rise to a bias in favour of those models that also make use of WordNet. To address this concern, we have created a new entailment dataset, \textbf{HP4K}, that makes use of Hearst Patterns, and is manually annotated, thereby avoiding the use of WordNet.

\textbf{Weeds:} The dataset is based on nouns sampled from WordNet where each noun had to occur at least $100$ times in Wikipedia, and its predominant sense had to account for more than 50\% of the occurrences in SemCor~\citep{Miller_1993}. We use the predefined split of~\citet{Weeds_2014b}, that avoids any lexical overlap between the training and evaluation sets. The split contains 2012 examples in the training set, evenly balanced between positive and negative hyponym-hypernym pairs, and 502 examples in the evaluation set.

\textbf{LEDS:} The dataset consists of 2770 examples, evenly balanced between positive and negative hyponym-hypernym pairs. The positive examples are based on direct hyponym-hypernym relationships from WordNet and the negative examples are based on a random permutation of the hypernyms of the positive pairs. As there is no predefined training/evaluation split, we make use of the 20-fold cross-validation methodology of~\citet{Roller_2016} that avoids any lexical overlap between training and evaluation sets.

\textbf{HP4K:} We extracted Hearst Patterns from our reference Wikipedia corpus and randomly selected 4500 unigram pairs. Subsequently, we manually annotated each pair according to whether it constitutes a correct hyponymy-hypernymy relation or not. The labelling was carried out by 4 experienced annotators --- all domain experts, familiar with the problem of hypernymy detection. We then split up the annotators in two teams, with each team annotating one half of the dataset. The initial round of annotations resulted in a Cohen's $\kappa$ score of 0.714, indicating substantial agreement~\citep{Viera_2005}. Conflicts were resolved on a cross-team basis such that team A would resolve team B's annotation conflicts and vice-versa. 

During annotation we noticed that positive pairs typically fall into one of two categories. Either they were ``true" subtype-supertype relations, such as $\langle\mbox{\em dog\/},\mbox{\em animal\/}\rangle$, or they were individual-class relationships where the hyponym is typically a named entity and represents a \emph{specific} instance of the more general class, as for example in $\langle\mbox{\em Nirvana\/},\mbox{\em band\/}\rangle$. Negative pairs were of a more diverse nature and included a range of different relations, such as co-hyponyms, meronyms or reverse hyponym-hypernyms. Negative pairs can also be comprised of two random nouns or two nouns without any semantic relation due to some amount of noise in extracting candidates solely on the basis of Hearst Patterns. Table~\ref{tbl:dataset_examples} shows positive and negative examples from the dataset.

\begin{table}[!htb]
\centering
\small
\resizebox{\columnwidth}{!}{
\begin{tabular}{ l l l }
\textbf{Pair}			& \textbf{Relationship}			& \textbf{Label} \\\hline
$\langle\mbox{\em dog\/},\mbox{\em animal\/}\rangle$		& hyponymy-hypernymy (Subtype)	& \emph{True}\\
$\langle\mbox{\em Nirvana\/},\mbox{\em band\/}\rangle$				& hyponymy-hypernymy (Instance)	& \emph{True}\\
$\langle\mbox{\em beef\/},\mbox{\em stew\/}\rangle$				& meronymy					& \emph{False}\\
$\langle\mbox{\em pedestrian\/},\mbox{\em road\/}\rangle$			& topical relatedness				& \emph{False}\\
$\langle\mbox{\em chemical\/},\mbox{\em adenosine\/}\rangle$		& reverse hyponymy-hypernymy	& \emph{False}\\
$\langle\mbox{\em cherry\/},\mbox{\em plum\/}\rangle$				& co-hyponymy					& \emph{False}\\
$\langle\mbox{\em form\/},\mbox{\em situation\/}\rangle$				& none						& \emph{False}\\\hline
\end{tabular}}
\captionsetup{font=small}
\caption{Examples from our proposed \textbf{HP4K} dataset.}
\label{tbl:dataset_examples}
\end{table}

\textbf{HP4K} consists of 4369 pairs with a class distribution of 45:55 ($\mbox{positive}:\mbox{negative}$). Subsequently we split the dataset into a training and evaluation set, ensuring that there is no lexical overlap between the two sets. This resulted in a training set of size 3426 and an evaluation set of size 943\footnote{All resources are available from \url{https://github.com/tttthomasssss/le-augmentation}.}. 

\subsection{Models}

We conduct experiments with two distributional vector space models, word2vec~\citep{Mikolov_2013b} and HyperVec~\citep{Nguyen_2017}. HyperVec is based on word2vec's skip-gram architecture and leverages WordNet to optimise the word representations for the hypernym detection task. Hierarchical information is encoded in the norm of the learned vectors, such that lexemes higher up in the hypernymy hierarchy have larger norms than lexemes in lower parts.

For word2vec we use the 300-dimensional pre-trained Google News vectors\footnote{Available from: \url{https://code.google.com/archive/p/word2vec/}.} and for HyperVec we trained 100-dimensional embeddings on a October 2013 Wikipedia dump~\citep{Wilson_2015}, using the recommended settings of~\citet{Nguyen_2017}, as our augmentation sets contained many words that were OOV in the pre-trained HyperVec vectors\footnote{We used the HyperVec code from \url{www.ims.uni- stuttgart.de/data/hypervec}.}.

In our experiments, we consider a supervised scenario where a classifier predicts a hyponym-hypernym relation between two given word embeddings. We use two different models as classifier a logistic regression classifier (LR), and a 3-layer feedforward neural network (FF). In both cases, the classifier takes the aggregated hypothesised hyponym-hypernym pair as input and predicts whether the pair is in a hyponym-hypernym relation. We report a detailed overview of the model parameterisation in Appendix~\ref{sec:supplemental_a}.

The two models share the same procedure for aggregating the word embeddings of the hypothesised hyponym-hypernym pair. For data augmentation based on distributional composition, we use vector averaging as composition function, which gave substantially better performance than addition in preliminary experiments.

\subsection{Results}

For the FF network, we performed 10-fold cross-validation on the \textbf{Weeds} and \textbf{HP4K} training sets. As our evaluation for \textbf{LEDS} is based on a 20-fold cross-valiation split, rather than a pre-defined training/evaluation split as for \textbf{Weeds} and \textbf{HP4K}, the same procedure for hyperparameter tuning is not straightforwardly applicable without exposing the model to some of the evaluation data. However, we found that the top parameterisations for \textbf{Weeds} and \textbf{HP4K} were quite similar and therefore applied hyperparameters to the FF model for \textbf{LEDS} that performed well in 10-fold cross-validation on \textbf{Weeds} \emph{and} \textbf{HP4K}. For data augmentation and dataset extension, we consider the following amounts of additional data: $\{0.2\text{K}, 1\text{K}, 2\text{K}, 4\text{K}, 10\text{K}, 20\text{K}, 40\text{K} \}$. All augmentation sets are balanced between positive and negative pairs.

\begin{figure*}[!htb]
\centering
\includegraphics[width=0.7\textwidth]{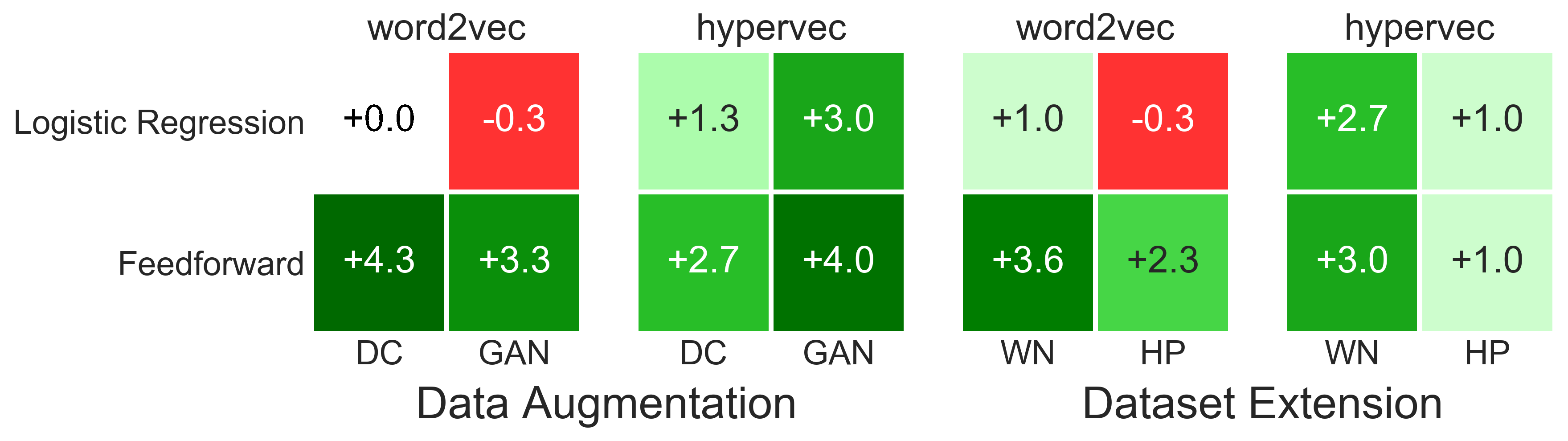}
\captionsetup{font=small}
\caption{Effect of data augmentation and dataset extension in absolute points of accuracy averaged across all datasets over the same model without augmentation or extension. The 2 heatmaps on the left are based on data augmentation (DC=Distributional Composition, GAN=\emph{GANDALF}). The 2 heatmaps on the right use dataset extension (WN=WordNet, HP=Hearst Patterns).}
\label{fig:avg_improvement}
\end{figure*}

Figure~\ref{fig:avg_improvement} shows the increase in absolute points of accuracy for the LR and FF model, as well as both vector spaces, averaged across all datasets. While in total data augmentation as well as dataset extension has a positive impact, the gains are larger for the FF model, suggesting that a higher capacity model is necessary to more effectively leverage the additional information from the augmentation source. Furthermore, before starting our experiments we exptected that extending an existing dataset with WordNet represents an upper bound on performance, given that WordNet is manually annotated and curated. However in our experiments we found that data augmentation by either distributional composition or by using \emph{GANDALF} remarkably \emph{surpassed} performance of the WordNet-based extension technique regularly. 


\begin{figure*}[!htb]
\centering
\includegraphics[width=0.7\textwidth]{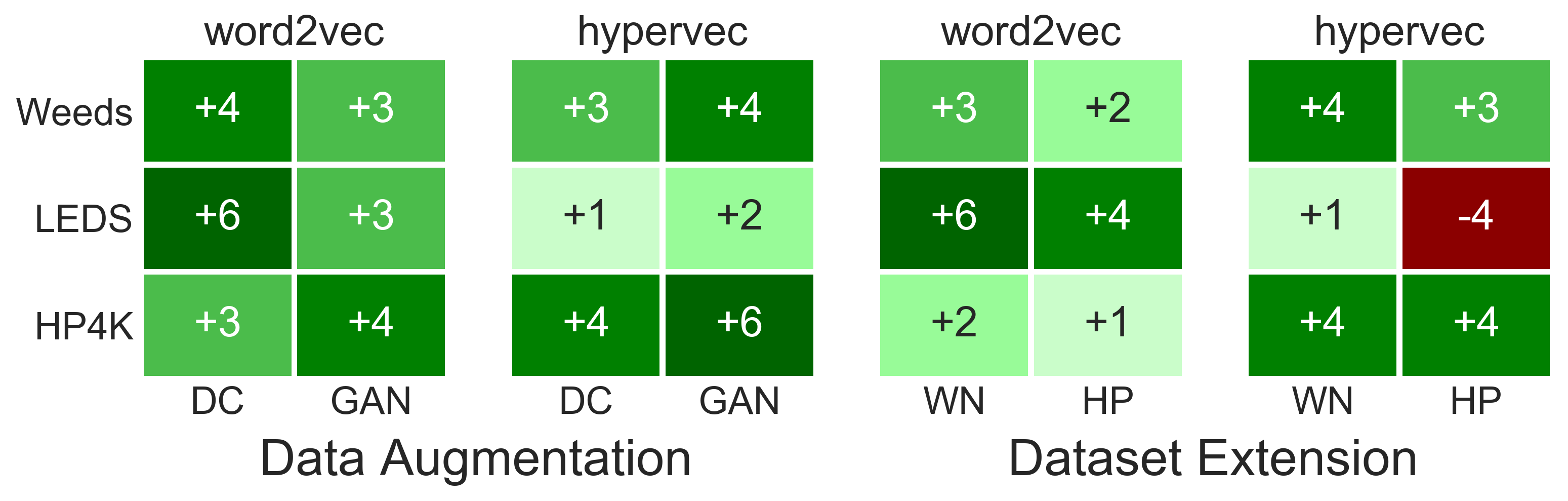}
\captionsetup{font=small}
\caption{Effect of data augmentation and dataset extension in absolute points of accuracy on all datasets for the FF model.}
\label{fig:ff_improvement}
\end{figure*}

The effect of data augmentation and dataset extension in absolute points of accuracy on each dataset individually for the FF model is shown in Figure~\ref{fig:ff_improvement}. It highlights consistent improvements across the board with only a single performance degradation in the case of extending the LEDS dataset with Hearst Patterns when using HyperVec-based word representations. The results per dataset for the LR model are presented in Appendix~\ref{sec:supplemental_a} and show that the LR model is less effective in leveraging the augmented data, causing more frequent performance drops. This suggests that models with more capacity are able to make more efficient use of additional data and are more robust in the presence of noise which is inevitably introduced by automatic methods.

Table~\ref{tbl:results} compares our FF model using word2vec embeddings with all proposed techniques for augmenting or extending a dataset. Our techniques are able to outperform a non-augmented model by 4-6 points in accuracy, representing a relative error reduction of  14\%-26\%. While the primary objective in this work is to improve an existing model setup with data augmentation, our augmented models compare favourably with previously published results.\footnote{We note that due to the use of different performance metrics and cross-validation splits, direct model-to-model comparisons are difficult on the LEDS and Weeds datasets. Thus we only compare to approaches that use the same evaluation protocol as we do.}
\begin{table}[!htb]
\centering
\small
\resizebox{\columnwidth}{!}{
\begin{tabular}{l | c | c | c}
\textbf{Model}								& \textbf{Weeds} 	& \textbf{LEDS}		& \textbf{HP4K} \\\hline
No Augmentation			& 0.72 			& 0.77			& 0.67 \\\hline 
Distributional Composition			& \textbf{0.76}		& \textbf{0.83}		& 0.70 \\
\emph{GANDALF} 				& 0.75			& 0.80			& \textbf{0.71} \\\hline
WordNet Extension							& 0.75			& \textbf{0.83}		& 0.69 \\
Hearst Patterns	 Extension						& 0.74			& 0.81			& 0.68 \\\hline\hline
\citet{Weeds_2014b}			& 0.75			& -				& - \\
\citet{Carmona_2017}	& 0.63			& 0.81			& - \\
\end{tabular}}
\captionsetup{font=small}
\caption{Accuracy scores for the data augmentation and the two dataset extension strategies in comparison to the same FF model without any augmentation or extension.}
\label{tbl:results}
\end{table}
In general, data augmentation by distributional composition or by \emph{GANDALF} overcomes two key weaknesses of simply extending a dataset with more data from WordNet or Hearst Patterns. First, many of the hyponym-hypernym pairs we mined from WordNet contain low-frequency words, which may have poor representations in our  vector space models. Second, while using Hearst Patterns typically returned higher frequency words, the retrieved candidates frequently did not represent hyponymy-hypernymy relationships. 


\section{Analysis}
\label{sec:analysis}
\begin{figure*}[!htb]
\centering
\setlength{\belowcaptionskip}{-10pt}
\setlength{\abovecaptionskip}{-10pt}
\includegraphics[width=\textwidth]{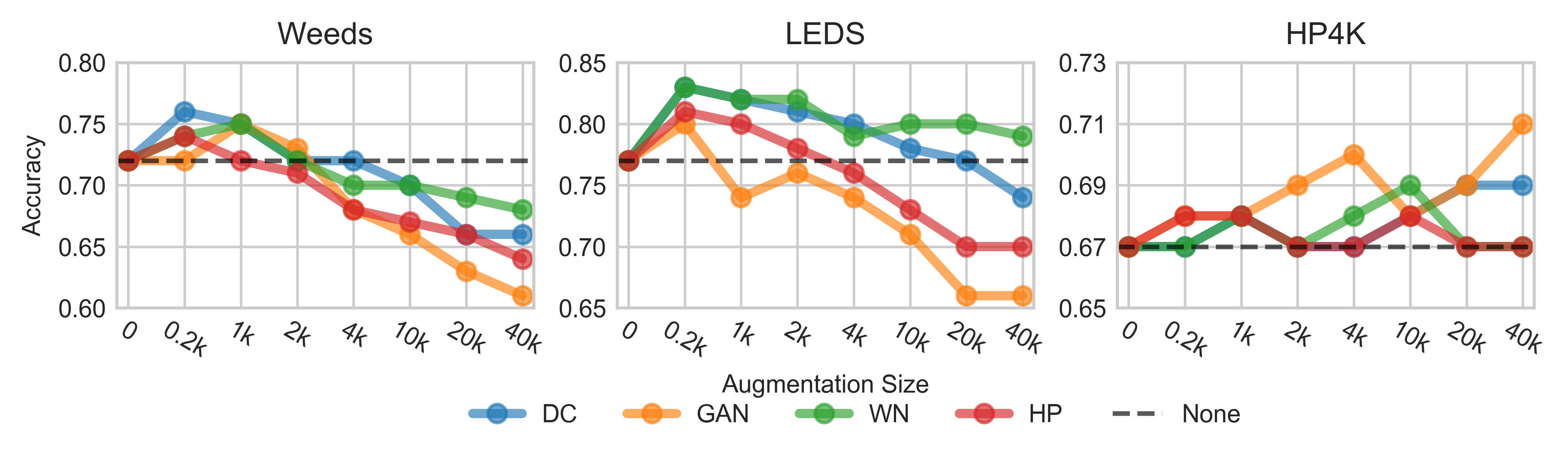}
\captionsetup{font=small}
\caption{Impact of different amounts of data augmentation.}
\label{fig:data_augmentation_size}
\end{figure*}

The concrete amount of data augmentation, i.e. the number of additional hyponym-hypernym pairs that are added to the training set, represents a tuneable parameter. Figure~\ref{fig:data_augmentation_size} shows the effect of varying amounts of data augmentation for the FF model, using word2vec representations, across all datasets. We note that all amounts of additional augmentation data share the same quality, i.e. it is not the case that a smaller augmentation set consists of ``better data" or contains less noise than a larger set.
For the \textbf{Weeds} and \textbf{LEDS} datasets, peak performance is typically achieved with smaller amounts of additional data, whereas for the \textbf{HP4K} dataset optimal performance is achieved with larger amounts of augmentation data. One explanation for the different augmentation characteristics of the \textbf{HP4K} dataset in comparison to the other two datasets is its independence of WordNet during the development of the dataset. 

\subsection{Data Augmentation in Space}

\begin{figure*}[!htb]
\centering
\setlength{\belowcaptionskip}{-2pt}
\setlength{\abovecaptionskip}{-2pt}
\includegraphics[width=0.7\textwidth]{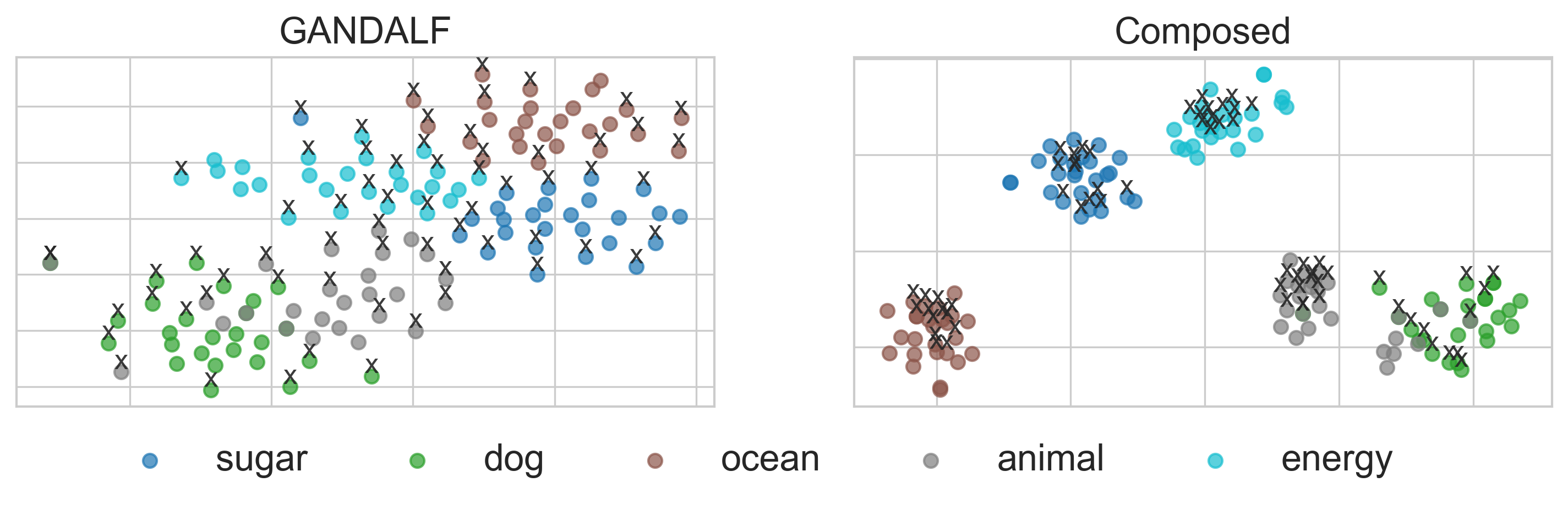}
\captionsetup{font=small}
\caption{t-SNE visualisation of the data augmentation spaces. Data points marked with ``x" denote the representation as coming from the data augmentation set.}
\label{fig:data_augmentation_spaces}
\end{figure*}

In order to visualise what area of the vector space the \emph{GANDALF} vectors and the composed vectors inhabit, we created a t-SNE~\citep{Maaten_2008} projection of the vector spaces in Figure~\ref{fig:data_augmentation_spaces}. For the visualisation we produced the nearest neighbours of standard word2vec embeddings and augmentation embeddings for 5 exemplary words and project them into the same space.  
Figure~\ref{fig:data_augmentation_spaces} shows that the generated augmentation points, marked with an ``x'', fit in with the real neighbours and do not deviate much from the ``natural" neighbourhood of a given word. \emph{GANDALF} vectors typically inhabit the edges of a neighbour cluster, whereas composed vectors are frequently closer to the cluster centroid. Table~\ref{tbl:augmentation_neighbours} lists the nearest neighbours for the example words. For word2vec and the composed augmentation vectors, we simply list the nearest neighbours of each query word. For \emph{GANDALF} we list the nearest neighbours of the generated vector that correspond to actual words. For example, if the vector \emph{GANDALF-234} is closest to the representations of \emph{sugar}, \emph{GANDALF-451} and \emph{mountain}, we only list \emph{sugar} and \emph{mountain} as neighbours of \emph{GANDALF-234}. 
\begin{table*}[!htb]
\centering
\small
\resizebox{\textwidth}{!}{
\begin{tabular}{ l | l | l | l}
\textbf{Word}			& \textbf{word2vec Neighbours}			& \textbf{\emph{GANDALF} Neighbours} & \textbf{Composed Neighbours} \\\hline
sugar			& refined sugar, cane sugar,    & cmh rawalpindi, prescribed antipsychotic medication,	& raw sugar, white sugar, brown sugar, \\
				& turbinado, cocoa, sugars	&  sugar, mumtaz bhutto, akpeteshie & sugar price, sugar industry\\\hline

dog				& dogs, puppy, pit bull, pooch, cat	& ellis burks, sniffing glue, microchip implants,  	& pet cat, dog fighting, cat breed, rat terrier\\
				& & liz klekamp, cf rocco baldelli & terrier breed\\\hline

ocean				& sea, oceans, pacific ocean, 	& pacific ocean, heavily vegetated, alaska aleutian, 				& ocean basin, pacific ocean, shallow sea, \\
					& atlantic ocean, oceanic & seagrasses, plutoid & sea fish, sea mammal\\\hline

animal				& animals, animal welfare, dog, pet, 	& animals, pet, hallway feeds, poop scooping, 		& first animal, adult animal, zoo animal, \\
					& cats & panhandle animal welfare & different animal, animal organization \\\hline

energy				& renewable energy, enery,  	& radial velocity measurements, stopped,  	& energy efficiency, solar energy, state energy, \\
					& electricity, enegy, fossil fuel & renewable energy, bicycle advisory, steinkuehler & food energy, energy company\\\hline

\end{tabular}}
\captionsetup{font=small}
\caption{Nearest neighbours for word2vec, GAN vectors and composed vectors.}
\label{tbl:augmentation_neighbours}
\end{table*}

The composed neighbours for each word are typically closely related to the original query, e.g. \emph{raw sugar} for \emph{sugar}, or \emph{zoo animal} for \emph{animal}. The \emph{GANDALF} neighbours on the other hand have a much weaker association with the query word, but are frequently still related to it \emph{somehow} as in the example of \emph{akpeteshie}, which is a spirit on sugar cane basis, as a neighbour for \emph{sugar}.

\subsection{Data Augmentation as Regularisation}

In the past, a prominent criticism of distributional methods for hypernymy detection was that such models were found to frequently identify features of a prototypical hypernym in the distributional representations, rather than being able to dynamically focus on the relevant features that are indicative of a hypernymy relation for a specific pair of words~\citep{Weeds_2014b,Levy_2015b}. We therefore briefly investigate whether data augmentation can be used as a regularisation mechanism that helps prevent models from overfitting on prototypical hypernym features. 

Table~\ref{tbl:results_hyper_only} shows the results on the \textbf{Weeds} dataset using a hypernym-only FF model with word2vec representations, in comparison to the same model variant that makes use of the hyponym and the hypernym. Ideally, we would hope to see weak performance for the hypernym-only and strong performance on the full model. This would indicate that the classifier does not rely on prototypical features in the hypernym, but is able to focus on specific features in a given hyponym-hypernym pair.
\begin{table}[!htb]
\centering
\small
\resizebox{\columnwidth}{!}{
\begin{tabular}{ l | l | l }
\textbf{Augmentation}	& \textbf{Hypernym-Only} & \textbf{Full}  \\\hline
No Data Augmentation				& 0.59		& 0.72 \\\hline
Distributional Composition (size=100)			& 0.60		& \textbf{0.74} \\
Distributional Composition (size=500)			& \textbf{0.57}	& 0.71 \\\hline
\emph{GANDALF} (size=500)		& \textbf{0.58}	& \textbf{0.75} \\
\emph{GANDALF} (size=1000)		& 0.60		& 0.73\\
\end{tabular}}
\captionsetup{font=small}
\caption{Accuracy for the hypernym-only and full models on the \textbf{Weeds} dataset with no, DC or GAN augmentation.}
\label{tbl:results_hyper_only}
\end{table}
For data augmentation by distributional composition there appears to be a correlation between the performance of the hypernym-only and the full model, i.e. a stronger model on the whole dataset also results in better performance for the hypernymy-only model. Hence augmentation by distributional composition might not be effective in helping the model to generalise in its current form. For augmentation with \emph{GANDALF} however, performance for the full model improves, while performance of the hypernym-only model slightly drops, suggesting that the evoked \emph{GANDALF} representations have a regularisation effect, while also improving generalisation. Hence, a fruitful avenue for future work will be further leveraging data augmentation for regularisation.

%



\section{Conclusion}
\label{sec:conclusion}
In NLP, in contrast to computer vision, data augmentation has not been applied as standard due to the apparent lack of universal rules for label-invariant language transformations.   

We have considered the problem of hypernymy detection, and proposed two novel techniques for data augmentation.   These techniques rely on semantic rules rather than an external knowledge source, and have the potential to  generate almost limitless synthetic data for this task.  We demonstrate that these techniques perform better than extending the training set with additional non-synthetic data, drawn from an external knowledge source in most cases. Our results are consistent across evaluation benchmarks, word vector spaces and classification architectures.  We have also shown that our approach is effective even when the word vector space model has already been specialised for hypernymy detection.

Since WordNet is widely used as a source of information about semantic relations, we have proposed a new evaluation benchmark that is  independent of WordNet.  Whilst results are lower across the board on this dataset, suggesting that it is more difficult than the others, we see the same pattern of increasing performance with a more complex classifier and the use of data augmentation. 

Future work includes leveraging data augmentation for more complex models and the extension of our approach to a multilingual setup as well as domains with a more specialised vocabulary such as Healthcare or Fashion.

\section*{Acknowledgements}
\label{sec:acknowledgements}
Part of this research was supported by EPSRC grant no. 2129720: \emph{Composition and Entailment in Distributed Word Representations}. We thank Mark Steedman for valuable feedback and insightful discussions that had substantial impact on the quality the paper.  We also thank several mini-batches\footnote{We used a mini-batch size of 3 for all of our submissions.} of reviewers for their \st{stochastic gradients} helpful comments that iteratively improved the paper, rejection after rejection, until it \emph{finally} converged towards acceptance.

\bibliography{common}
\bibliographystyle{acl_natbib}

\clearpage
\appendix

\section{Supplemental Material}
\label{sec:supplemental_a}
\subsection{GANDALF Model Details}

The generator and discriminator in \emph{GANDALF} are single layer feedforward networks, with tanh activations and a dropout ratio~\citep{Srivastava_2014} of 0.3. We used ADAM~\citep{Kingma_2014} to optimise a binary cross entropy error criterion with a learning rate of 0.0002 and $\beta$ values of 0.5 and 0.999. We found that \emph{GANDALF} required quite a bit of wizardry to achieve strong performance and we found the website \url{https://github.com/soumith/ganhacks} very helpful. For example we applied label noise and soft labels~\citep{Salimans_2016} and used a batch normalisation layer~\citep{Ioffe_2015}, which had the largest impact on model performance. \emph{GANDALF} is implemented in PyTorch~\citep{Paszke_2017} and we release our code on \url{website}.

\subsection{Model Details}

For our linear model we use the logistic regrssion classifier implemented in scikit-learn~\citep{Pedregosa_2011}. Our neural network model is 3-layer feedforward model implemented in PyTorch~\citep{Paszke_2017}.

We tuned the parameters of the Feedforward neural network by 10-fold cross-validation on the respective training sets, except for \textbf{LEDS}, where we chose the parameters on the basis of a model that performed well on the \textbf{Weeds} and \textbf{HP4K}. Our parameter grid consisted of activation function: \{tanh, relu\}, dropout: \{ 0.0, 0.1, 0.3 \} and hidden layer sizes, where we considered \{ 200-200-200, 200-100-50, 200-50-30 \} for Hypervec and \{ 600-600-600, 600-400-200, 600-300-100, 600-200-50\} for word2vec. We furthermore considered 3 different aggregation functions: \texttt{diff}~\citep{Weeds_2014b}, which simply takes the elementwise difference of the embedding pair; \texttt{asym}~\citep{Roller_2014} which is the concatenation of the difference and the squared difference of the embedding pair; and \texttt{concat-asym}~\citep{Roller_2016}, which is the concatenation of the embedding pair, their difference, and their squared difference. We trained all models for 30 epochs with early stopping and used ADAM with a learning rate of 0.01 to optimise a cross entropy error criterion. 

\subsection{Results}

Figure~\ref{fig:lr_improvement} below shows the effect of data augmentiation in terms of points of Accuracy for the logistic regression classifier per vector space model and dataset. Unlike for the higher-capacity feedforward model, data augmentation frequently causes performance to go down for the simpler linear model. This suggests that more complex models are required to fully leverage the additional information from the augmentation sets.
\begin{figure*}[!htb]
\centering
\includegraphics[width=0.7\textwidth]{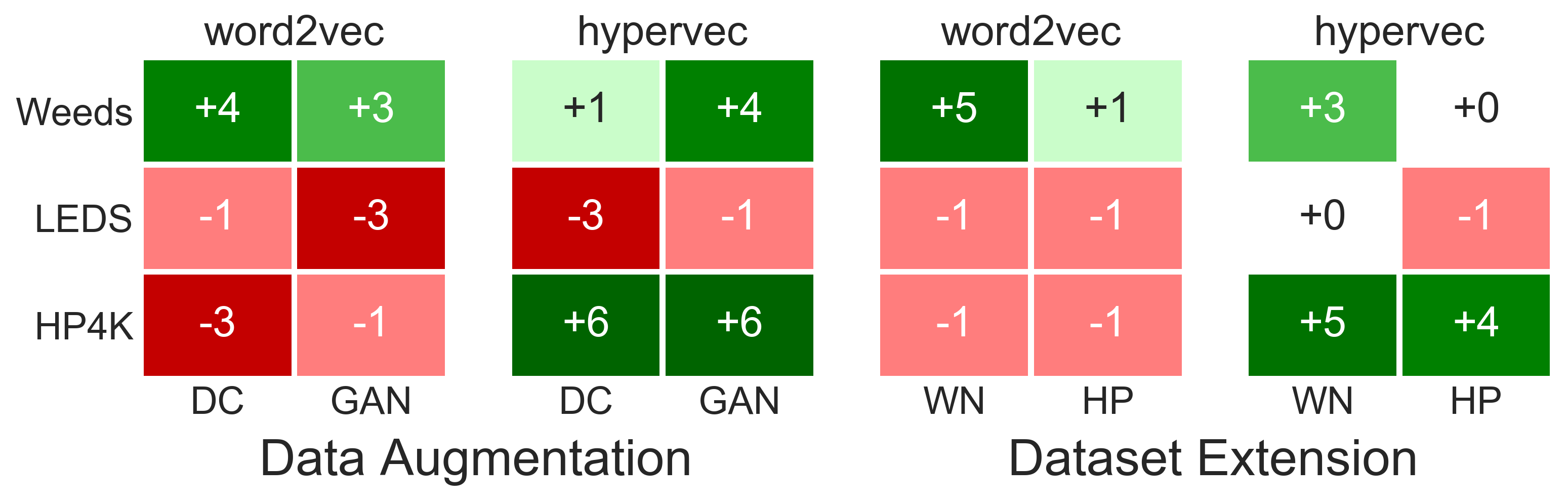}
\captionsetup{font=small}
\caption{Effect of data augmentation and dataset extension in absolute points of accuracy on all datasets for the LR model.}
\label{fig:lr_improvement}
\end{figure*}

Table~\ref{tbl:all_results} below gives an overview over the complete results for both classifier and vector space models, across all datasets.
\begin{table*}[!htb]
\centering
\small
\resizebox{\textwidth}{!}{
\begin{tabular}{ l | c c c c c | c c c c c | c c c c c | l}
	& \multicolumn{5}{c}{\textbf{Weeds}}		&	\multicolumn{5}{c}{\textbf{LEDS}}	& \multicolumn{5}{c}{\textbf{HP4K}}	&  \\
\textbf{Model}	& \textbf{None} & \textbf{DC} & \textbf{GAN} & \textbf{WN} & \textbf{HP} & 		\textbf{None} & \textbf{DC} & \textbf{GAN} & \textbf{WN} & \textbf{HP} & \textbf{None} & \textbf{DC} & \textbf{GAN} & \textbf{WN} & \textbf{HP} & \textbf{Vector Space} \\\hline
LR			& 0.69 & \underline{0.73} & \underline{0.72} & \underline{0.74} & \underline{0.70}				& 0.81 & 0.80 & 0.78 & 0.80 & 0.80													& 0.64 & 0.61 & 0.63 & 0.63 & 0.63			& \multirow{2}{*}{word2vec}\\
FF			& 0.72 & \textbf{\underline{0.76}} & \underline{0.75} & \underline{0.75} & \underline{0.74}			& 0.77 & \textbf{\underline{0.83}} & \underline{0.80} & \textbf{\underline{0.83}} & \underline{0.81}	& 0.67 & \underline{0.70} & \textbf{\underline{0.71}} & \underline{0.69} & \underline{0.68}	&\\\hline
LR			& 0.70 & \underline{0.71} & \underline{0.74} & \underline{0.73} & 0.70						& 0.79 & 0.76 & 0.78 & 0.79 & 0.78													& 0.63 & \underline{0.69} & \underline{0.69} & \underline{0.68} & \underline{0.67}			& \multirow{2}{*}{hypervec}\\
FF			& 0.71 & \underline{0.74} & \textbf{\underline{0.75}} & \textbf{\underline{0.75}} & \underline{0.74}	& 0.79 & \underline{0.80} & \textbf{\underline{0.81}} & \underline{0.80} & 0.75					& 0.66 & \underline{0.70} & \textbf{\underline{0.72}} & \underline{0.70} & \underline{0.70}	&\\
\end{tabular}}
\captionsetup{font=small}
\caption{Accuracy scores for the data augmentation strategies (DC and GAN), and the two dataset extension strategies (WN and HP), and the baseline that neither uses augmentation nor extension (None). Boldfaced results denote top performance per vector space and dataset, underlined results denote improved performance in comparison to the baseline without data augmentation.}
\label{tbl:all_results}
\end{table*}
It shows the consistent positive effect of data augmentation on the more complex feedforward model in comparison to the logistic regression classifier, which is less robust to the small amounts of noise that is inevitably introduced by the automatic augmentation algorithm. 
\end{document}